\documentclass[conference]{IEEEtran}
\usepackage{amsmath,amsfonts}
\usepackage{algorithmic}
\usepackage{algorithm}
\usepackage{array}
\usepackage[caption=false,font=normalsize,labelfont=sf,textfont=sf]{subfig}
\usepackage{textcomp}
\usepackage{stfloats}
\usepackage{url}
\usepackage{verbatim}
\usepackage{graphicx}
\usepackage{cite}
\usepackage{bm}
\usepackage{tabularx}
\usepackage{multirow}
\usepackage{tabu}
\usepackage{tabularx,ragged2e}
\usepackage{booktabs}
\usepackage{subcaption}
\usepackage{graphicx}     
\usepackage{subcaption}   
\begin{document}

\title{Characterizing the Behavior of Training Mamba-based State Space Models on GPUs}



\author{%
\IEEEauthorblockN{Trinayan Baruah\IEEEauthorrefmark{1} \ \ \ Kaustubh Shivdikar\IEEEauthorrefmark{1} \ \ \ Sara Prescott\IEEEauthorrefmark{2} \ \ \ David Kaeli\IEEEauthorrefmark{3}}
\IEEEauthorblockA{\IEEEauthorrefmark{1}Advanced Micro Devices (AMD), \texttt{\{tbaruah,kshivdik\}@amd.com}}
\IEEEauthorblockA{\IEEEauthorrefmark{2}Massachusetts Institute of Technology (MIT), \texttt{sprescot@mit.edu}}
\IEEEauthorblockA{\IEEEauthorrefmark{3}Northeastern University (NEU), \texttt{d.kaeli@northeastern.edu}}
}




\maketitle

\begin{abstract}
Mamba-based State Space Models~(SSM) have emerged as a promising alternative to the ubiquitous transformers. Despite the expressive power of transformers, the quadratic complexity of computing attention is a major impediment to scaling performance as we increase the sequence length. SSMs provide an alternative path that addresses this problem, reducing the computational complexity requirements of self-attention with novel model architectures for different domains and fields such as video, text generation and graphs. Thus, it is important to characterize the behavior of these emerging workloads on GPUs and understand their requirements during GPU microarchitectural design. In this work we evaluate Mamba-based SSMs and characterize their behavior during training on GPUs. We construct a workload suite that offers representative models that span different model architectures. We then use this suite to analyze the architectural implications of running Mamba-based SSMs on GPUs. Our work sheds new light on potential optimizations to continue scaling the performance for such models.
\end{abstract}


\section{Introduction}
The past few years have seen  explosive growth in the field of Deep Learning~(DL). Early DL models were primarily used for tasks such as product recommendations, search and social media recommendations.  In contrast, DL models today directly interact with the users via services such as chatbots, coding-assistants and video generation. This major transformation over the past few years has been driven by the development of Generative AI models, such as Large Language Models (LLMs) and vision-based LLMs. 

Despite the advances made in recent years by LLM, scaling training of these models has become a major challenge due to the performance of the self-attention layer~\cite{dao2023flashattention}. One of the primary challenges in scaling LLMs for training is the quadratic dependence of the compute time on the sequence length. The problem is further exacerbated by the fact that the softmax operation is usually done at higher precisions (e.g., FP16 and FP32) using the vector and exponential units on the GPUs. Today's GPUs are optimized to run matrix-multiplication~(GEMM) on tensor cores, which are optimized for computing GEMMs using lower precision such (e.g., FP4 and FP8). If we consider the FP8 tensor compute throughput on current GPU, such as NVIDIA Hopper H100, the ratio between that and the exponential throughputs is ~512. As identified by Shah et al.~\cite{shah2024flashattention}, the softmax operator is a major bottleneck and limits efficient performance scaling as a function of sequence length. Machine learning researchers have come up with different methods to address this issue. One approach is to focus only on a given window of tokens~\cite{beltagy2020longformer},  as used in models such as Mistral7B and LLAMA4. The other approach is to develop foundational models that overcome the quadratic computational scaling barriers, as offered by State Space Models~(SSMs), which are the subject of this paper.


\begin{figure}[!t] \centering
\includegraphics[width=\linewidth]{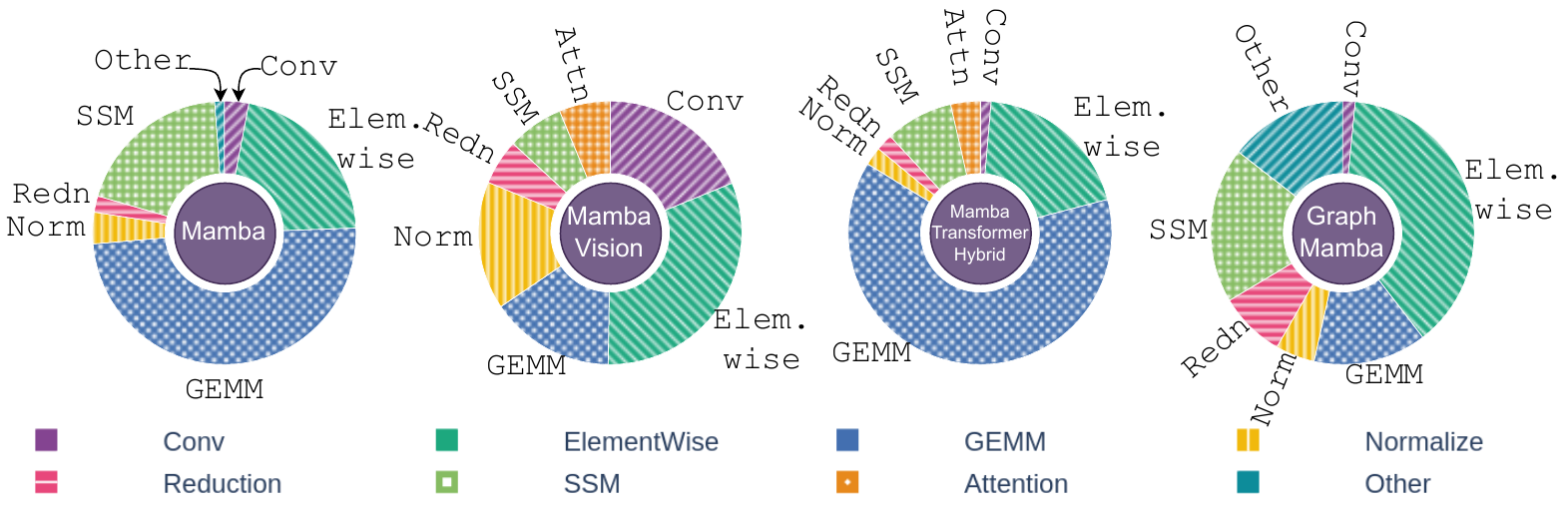} \caption{Mamba Operator Breakdown\label{fig:operatorbreakdown}} 
\end{figure}

SSMs have emerged as a promising alternative to tackle this problem in recent years. The core idea is borrowed from Recurrent Neural Networks~(RNNs) where the output only depends on the previous token instead of attention, where for each output token the model needs to consider all previous tokens. SSMs instead have a hidden state that holds a compressed history of all previous tokens. The sequential nature of the original SSMs, such as S4~\cite{gu2021efficiently},  impact their scalability on parallel hardware such as a GPU. Additionally, the original SSM models did not perform well on various tasks such as copying and induction heads. Mamba~\cite{gu2023mamba} , ~\cite{dao2024transformers} was proposed as a selective SSM which was designed to solve both issues. As a result, Mamba has been adopted in popular libraries such as NVIDIA-Megatron~\cite{narayanan2021efficient}.
Given the ubiquity of GPUs for training and inference for such Mamba based models, we need to better understand the performance implications of training Mamba-based models. If we can characterize the common patterns and operators present in SSM models, this information can help guide future hardware design. Identifying architecture bottlenecks present on the GPU when running SSMs can help guide software developers to better optimize their code. In this paper we characterize the behavior of Mamba-based SSM models on GPUs. We provide detailed characteristics of the key SSM operators in Mamba, dissecting key computational and memory behavior, as well as identifying the key bottlenecks so that we can effectively scale the performance of such models on future systems.

\vspace{-0.5em}
\section{Background of SSMs and Mamba}
\vspace{-0.5em}
In this section, we describe of State Space Models (SSMs), which serve as the
foundational building block of Mamba. 
We refer readers to the work of Gu et al.~\cite{gu2021efficiently,gu2023mamba} and Dao et al.~\cite{dao2024transformers} for a deeper dive into the mathematics underlying these models. Fundamentally, SSMs can be defined using four matrices \( A, B, C, D \), a hidden state \( h \), an input \( x \), and an output \( y \). The general formulation of an SSM is given by Equations ~\eqref{eq:ssm_state} and~\eqref{eq:ssm_output}.

\begin{align}
x_{k+1} &= A x_k + B u_k  \label{eq:ssm_state}\\
y_k &= C x_k + D u_k \label{eq:ssm_output}
\end{align}

The state equation of an SSM captures how the hidden state is updated using matrix A, as well as the influence of the input \textit{x} on matrix B. The output equation captures how the output evolves based on the impact of the state with matrix C, and includes the impact of the input with matrix D. In SSMs such as S4~\cite{gu2021efficiently}, the state-space matrices are independent of the input, which makes them ineffective for certain tasks such as  selective copying and induction heads~\cite{gu2023mamba}.

To address these limitations, Mamba~\cite{gu2023mamba} introduced  modifications in how the state matrices are handled. While the \( A \) matrix remains constant, the \( B \) and \( C \) matrices are now dynamically computed based on the input, making the model content-aware. To enable efficient execution on parallel hardware, such as GPUs, Mamba employs a parallel scan algorithm~\cite{blelloch2002scans}. Modern GPUs are heavily optimized for matrix multiplication, with specialized acceleration capabilities such as tensor cores. To further accelerate computation, the developers of Mamba have proposed imposing structural constraints on the \( A \) matrix so that the operations can be reformulated as a matrix-multiplication algorithm~\cite{dao2024transformers} instead of using a parallel scan. In this paper, we refer to this implementation as Mamba2. In a Mamba-based SSM model, three of the matrices (namely,  \textbf{X}, \textbf{B} and \textbf{C}) undergo an input projection layer which is basically a fully-connected layer.  This is then followed by a one-dimensional convolution and a nonlinear activation function~(typically a softplus) applied to the matrices.   Matrix A is passed through the input projection layer without going through a convolution and non-linear activation layer. All four matrices are then fed into the SSM block, which is the core innovation of Mamba's based models.  This step performs the operations shown in Equations ~\eqref{eq:ssm_state} and~\eqref{eq:ssm_output}.

\section{Workload Selection and Evaluation}

Our goal in this work is twofold. We want to understand the variations at the machine learning model level for Mamba-based SSM models. We also want to understand the behavior and architectural implications of the SSM block, which is the key block in such models.  We also would like to identify how Mamba execution is different than other ML models such as Convolutional Neural Networks~(CNNs) and Large Language Models~(LLMs). 

Surveying the range of SSM models developed, we find that the number of Mamba-based SSM models is rapidly increasing, though they all share certain common characteristics. While the original Mamba models~\cite{gu2023mamba} consists of stacking Mamba-only blocks as multiple layers, there are also approaches that combine attention blocks from the transformers~\cite{vaswani2017attention} along with Mamba blocks, stacking them in a desired ratio. For example, the best performing model in terms of accuracy~\cite{waleffe2024empirical} has 7\% of the layers as attention layers~\cite{vaswani2017attention} and the rest of the layers as Mamba layers.

We find differences between model architectures that are task-specific. For example, language modeling, image processing and graph processing require inherently different model architectures due to the variation in the type of inputs they process. This is because language modeling data is a one-dimensional representation of words, whereas image and video data represent a two-dimensional representation, as well as having spatial and temporal properties. In contrast, graph data is inherently non-euclidean~\cite{zhou2020graph}. We take these factors into account and come up with a suite of four different models, as described below.

\begin{itemize}
\item \textbf{Homogeneous Mamba Model}: A homogeneous Mamba models, such as the ones introduced by Dao et al.~\cite{dao2024transformers} and waleffe et al.~\cite{waleffe2024empirical}, consist of multiple Mamba layers, one after the other, and are developed for language modeling tasks. The key block in such model architectures consists of either a Mamba-1 or Mamba-2 block, which performs input and output projection operations~(typically matrix multiplication), a one dimensional causal-convolution, an activation function and the key SSM operation. We use the open-source implementation that is part of NVIDIA's Megatron library~\cite{narayanan2021efficient} as a representative workload for a homogeneous Mamba model.  We use the open-source bookcorpus dataset as input to our study. The model adopts the Mamba-2 algorithm given the ability of Mamba-2 to exercise the tensor computation on GPUs (unlike Mamba-1). 

\item \textbf{MambaVision}: MambaVision~\cite{hatamizadeh2025mambavision} extends the original Mamba model~\cite{gu2023mamba} to vision tasks by introducing a novel model architecture that is comprised of a mixture of different operators, namely convolution, a Mamba block and a self-attention block.
This Maba model is used for image-based tasks. Unlike the one-dimensional causal convolution in the language-based Mamba models, Mamba vision uses a standard non-causal convolution, which is commonly used in image processing tasks. The modified Mamba layers, as well as the standard self-attention layers, are followed by multi-layer perceptron~(MLP) layers. We use the open-source implementation of this model and use the Imagenet~\cite{deng2009imagenet} dataset for our analysis.

\item \textbf{Mamba-Transformer Hybrid}: Recent research~\cite{waleffe2024empirical,lieber2024jamba} has shown that combining Mamba, self-attention and multi-layer perceptron~(MLP) blocks provide better results in terms of accuracy than using Mamba-only layers. This is because self-attention can capture local attention dynamics more precisely, whereas the Mamba blocks can capture global dynamics, while providing the ability to scale linearly in terms of computational complexity with sequence length. The model architecture consists of Mamba and self-attention layers. We use the implementation from NVIDIA's Megatron library and build a representative model with 50\% MLP Layers and 7.1\% of self-attention layers and the remaining 43\% as Mamba-2 layers, similar to the model described by Waleffe et al.~\cite{waleffe2024empirical}, and run the model using the bookcorpus dataset.

\item \textbf{GraphMamba}: GraphMamba~\cite{behrouz2024graph} is an approach proposed to extend the Mamba architecture model to support graphs, developing a Graph Mamba block~(GMB). The model is a variant of Graph Neural Networks~(GNNs)~\cite{zhou2020graph} which have been used heavily in graph processing tasks.   The node embeddings are passed through the GMB block, whereas the edge embeddings pass through the message passing neural network, which is a staple of all Graph Neural Networks~\cite{zhou2020graph}. Thus, this model represents a hybrid architecture using blocks from both GNNs and Mamba.
\end{itemize}

\section{Evaluation}
We evaluate these workloads on an NVIDIA Hopper H100  SXM architecture available on the JarvisLabs cloud~(specifiations listed in Table ~\ref{tab:h100}. We enable flash attention optimizations~\cite{dao2023flashattention}, so that for models that include self-attention, we can accelerate the self-attention computation. We profile a full training loop iteration over a minibatch, but exclude the first iteration since that usually involves lot of automated library tuning to select the most performant kernels for a given input. We profile specific iterations by instrumenting the code using NVTX, and leverage start and stop markers to ensure the profiling captures the relevant iteration for both the forward and backward passes of the model.  In our study, we also conduct an in-depth characterization of the key SSM block kernels. For this, we use the parameters for the SSM block from prior work~\cite{waleffe2024empirical}.

\begin{table}[h]
  \centering
  \begin{tabular}{|c|c|}
    \hline
    Category & Value\\
    \hline
    Num SMs & 132 \\
    \hline
    BFP16 Tensor TFLOPs, FP32 Tensor TFLOPs & 989,494\\
    \hline
    FP32 Vector TFLOPS & 133.8 \\
    \hline
    Memory type and capacity  & HBM3,80GB\\
    \hline
    Memory bandwidth & 3350 GB/s \\
    \hline
    L2 Cache Size, L1 Cache Size & 50 MB,256 KB \\
    \hline
    
  \end{tabular}
  \caption{H100 Hardware Parameters}
  \label{tab:h100}
\end{table}

\section{Results}
In this section we present our workload  characterization of Mamba-based models during training. Evaluation is run on a single GPU.   Our goal is to capture profiles that allow us to quantify arithmetic intensity, breakdown of different operators within a model, as well as understand memory behavior. We also reflect on the architectural implications of the SSM kernels when mapped to a GPU. 
\subsection{Operator breakdown}
Figure~\ref{fig:operatorbreakdown} shows an operator-level breakdown for Mamba based models. While matrix-multiplication operations~(GEMMs) are dominant~(up to 60\% of total execution),  SSM operators also occupy a significant portion of the total execution~(up to 20\%). The amount of time spent in SSM-related kernels is lower in hybrid Mamba transformer models, as well as in vision models, since other operators, such as attention and convolution, are also present. Thus, we need to understand and characterize this behavior so that we can understand and improve the SSM block performance for future GPUs. Apart from the other operators (e.g., element-wise and normalization operators), a significant portion of the execution is dominated by Mamba based models~(up to 30\% of the total time). This suggests that scaling the performance of such models goes beyond simply scaling the tensor arithmetic FLOPS of GPUs, as element-wise and normalization operators tend to be memory bandwidth bound.
\subsection{Arithmetic Intensity of the Network}
\begin{figure}[!t] \centering
\includegraphics[width=\linewidth]{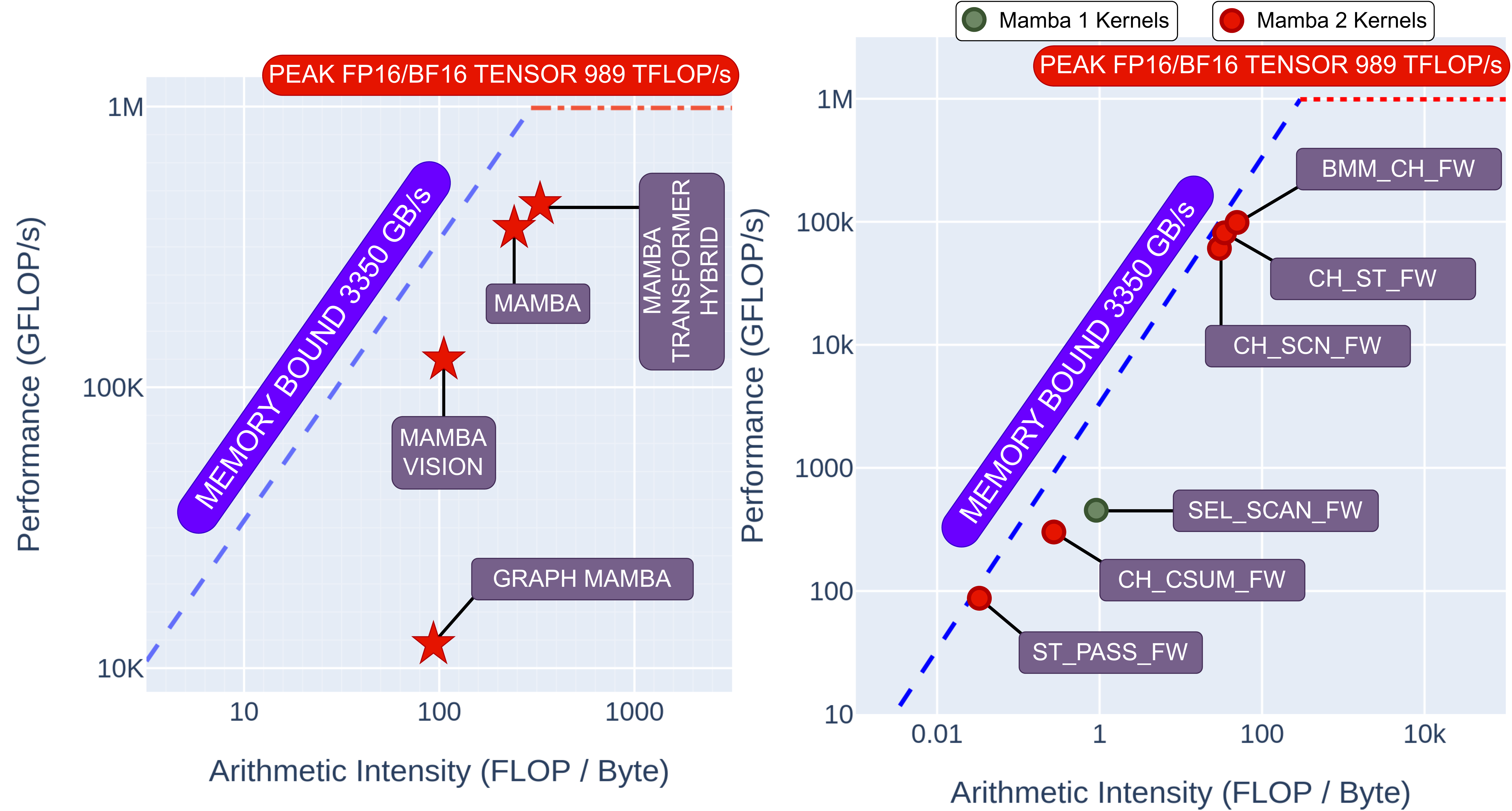} \caption{Roofline of the key Mamba SSM kernels in forward and backward passes .\label{fig:kernel_roofline}} 
\end{figure}

Figure~\ref{fig:kernel_roofline} shows the arithmetic intensity of the different Mamba based models (left), as well as the arithmetic intensity of the SSM kernels (right). For the kernel roofline plot (right), we only include the kernels in the forward pass for clarity, though the same observations apply for the backward pass.

As we can see, all the models are memory bound, suggesting that this is the primary bottleneck. In general, models with highest number of GEMM operators~(62\% in Mmaba Transformer Hybrid) are more compute bound, versus models that have fewer operators (e.g., Graph-Mamba only spends 14\% of time spent in GEMM). Other operators, such as element-wise operators, tend to be memory bound.

When we look at the arithmetic intensity of the SSM block kernels, we observe that they also have low arithmetic intensity and thus are not compute bound. However, some kernels, such as bmm\_ch\_fwd, do have a higher arithmetic intensity compared to others, such as state\_pass\_fwd. As we will see later, these higher arithmetic intensity kernels are the ones that leverage tensor compute. Mamba1 has only one kernel in the forward pass~(i.e., sel\_scn\_fwd) and the arithmetic intensity of this kernel is much lower than some of the kernels from Mamba2. In terms of hardware implications, increasing the tensor compute FLOPS will only accelerate the GEMM operations in  these networks. Exploring hardware and software techniques to better manage memory bandwidth will be needed in order to continue scale the performance of such models. We will provide more detailed insights into the compute  and memory behavior of the SSM block kernels in the upcoming sections.


\begin{figure}[!t] \centering
\includegraphics[width=\linewidth]{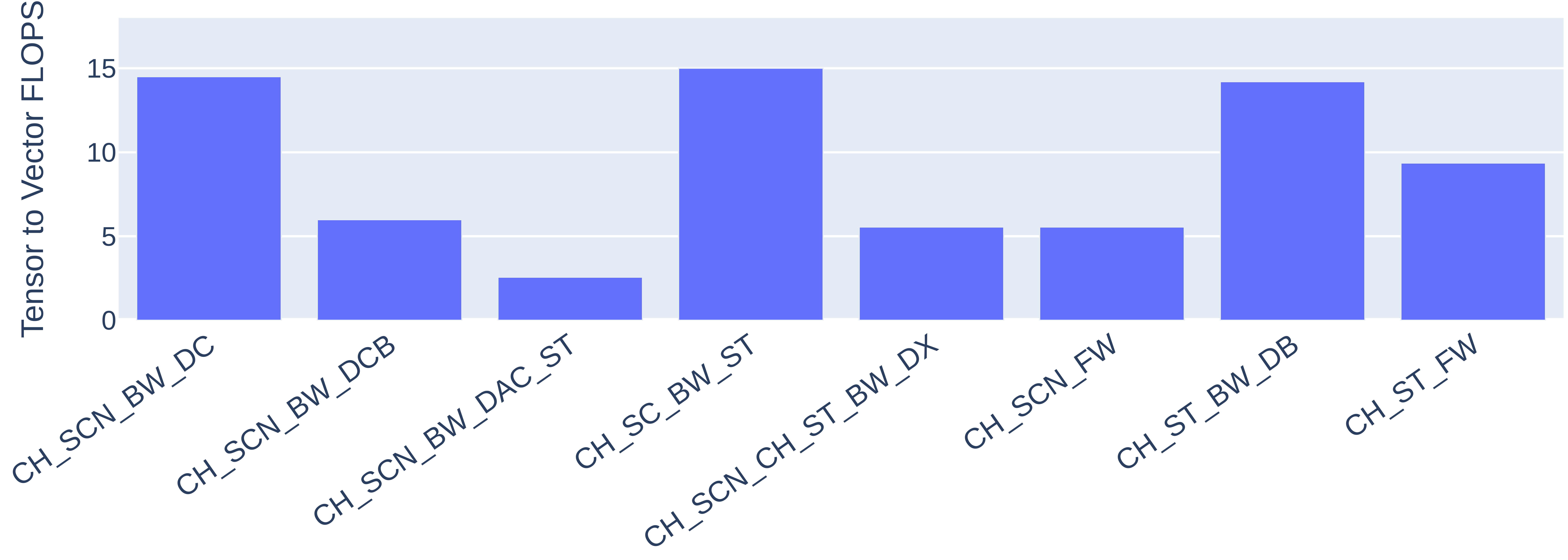} \caption{Tensor to Vector FLOPS Ratio.\label{fig:tensor2vector}} 
\end{figure}



\subsection{Memory and Cache behavior of SSM kernels}
\begin{figure}[!t] \centering
\includegraphics[width=\linewidth]{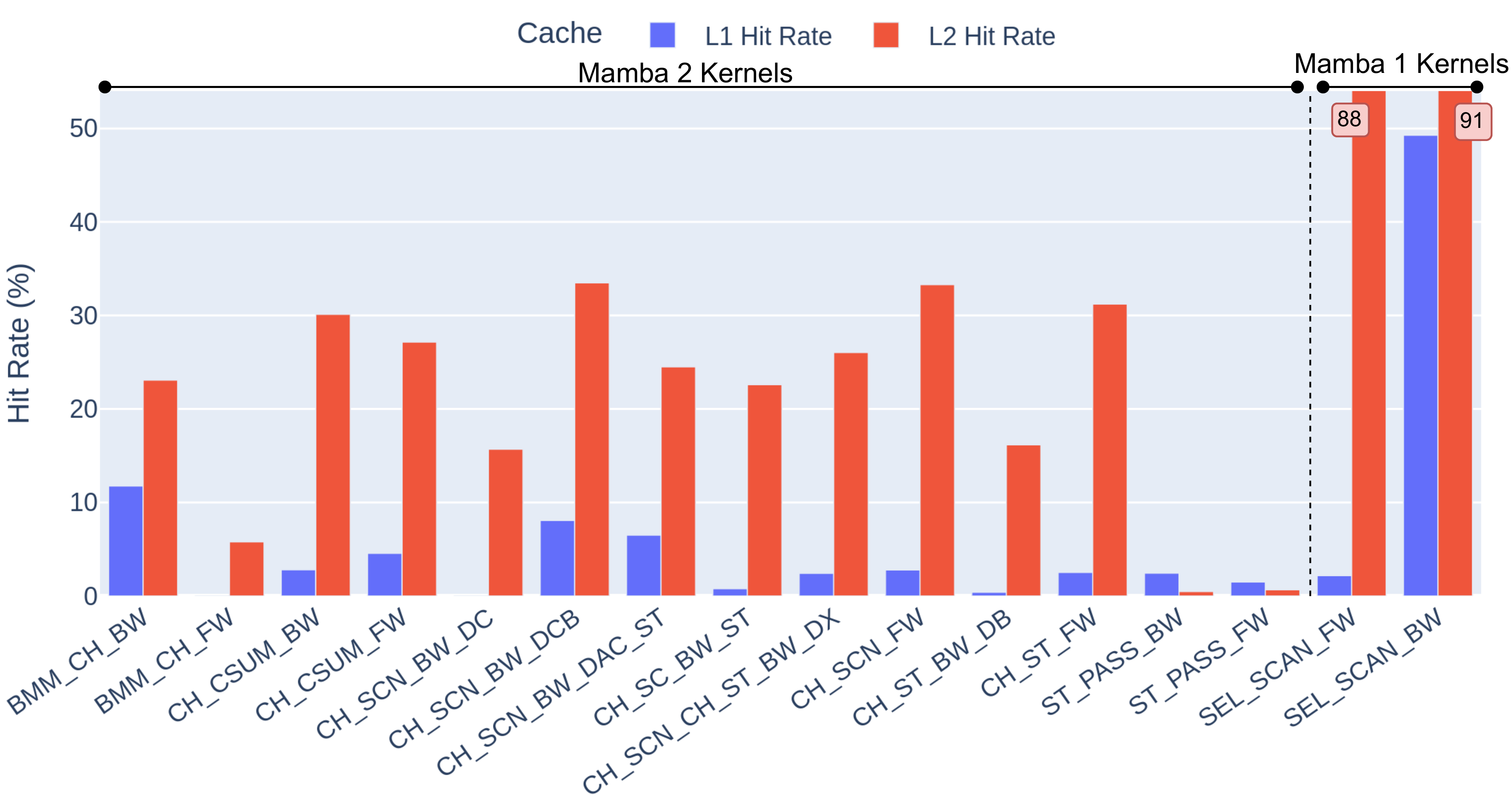} \caption{L1 and L2 cache hit rate for SSM kernels.\label{fig:hitrate}} 
\end{figure}

Figure ~\ref{fig:hitrate} shows the hit rates for the SSM block kernels for both the L1 and the L2 cache on the H100 GPU. We find that the data access patterns for these kernels exhibit two different types of locality. For kernels with tensor operations, there is a degree of spatial locality across the thread-blocks running on different GPU SMs, as is typical in any GEMM-based tiled kernel. There is also locality across kernels since the output of one kernel becomes the input of a subsequent kernel showing a typical write to read dependency chain. For example, the kernel bmm\_ch\_fwd computes the dot product of B and C tensors in a tiled fashion and stores the output in global memory. The next kernel that is executed (i.e., ch\_scan\_fwd) uses this as an input during its execution and as a result can get the locality benefits of L2 cache during execution provided the writes from bmm\_ch\_fwd can fit in the cache. Such dependencies across kernels also show potential kernel fusion optimizations that can be leveraged to avoid even having writing to the L2 cache in the first place. The L1 cache hit rate is low primarily for two reasons. First, all kernels that use tensor compute read the data from global memory to shared memory,  to stage their computation there as is typical in GEMM and GEMM like operations such as attention. And other kernels that do not use tensor computations, such as state\_pass\_fwd, do not typically have any significant data reuse potential within the SM to get benefits of L1 caching. In terms of memory bandwidth utilization, despite the SSM kernels inherently having a low arithmetic intensity, and thus are memory bound, we observe they are not able to fully saturate the available memory bandwidth available, which is 3350 GB/s. Our analysis shows that, on average, the utilization is only 57.5\% from the peak. This is because we observe that current implementations of the key kernels such as ch\_scan\_fwd  produce thread blocks with very short lifetime. The dot products in these kernels involve a very small contraction dimension~(typically referred as the GEMM K) since the contraction dimension is either the chunk size or the head dimension for these kernels. Thus, the actual useful work done inside the thread block is very short as opposed to the launch, setup and teardown overheads of a thread block. Software optimizations, such as persistent thread-blocks~\cite{gupta2012study}, which have also been replaced with dedicated hardware support on newer GPUs such as NVIDIA Blackwell B200, can help alleviate these overheads so that the kernels can saturate the available memory bandwidth more effectively.

\begin{table}[h]
  \centering
  \begin{tabular}{|c|c|c|c|}
    \hline
    Category & TensorOnly & VectorOnly & Tensor+Vector \\
    \hline
    Forward & 1 kernel, 2.40\% & 2 kernels, 8.55\% & 2 kernels,89.03\% \\
    \hline
    Backward & 1 kernel, 1.26\% & 2 kernels, 6.86\% & 6 kernels,91.87\% \\
    \hline
  \end{tabular}
  \caption{Breakdown of Mamba-2 kernels using tensor only, tensor only and both tensor+vector along with the percentage of their runtime within the SSM block for forward and backward pass respectively}
  \label{tab:tensorvectortable}
\end{table}

\subsection{Tensor to Vector Ratio of Mamba2 kernels}
In our experiments, we observe that Mamba2 has kernels that exercise both the tensor and vector compute. Table ~\ref{tab:tensorvectortable} shows the breakdown of the number of kernels that use either the tensor units, the vector units and those that use both the tensor and vector units in the forward and backward pass respectively for Mamba2. We also record the percentage of time taken for these separate types of kernels during execution for the representative parameters we use in our study~\cite{waleffe2024empirical} . The key observation is that a significant number of kernels, specifically two in the forward, pass out of five and six in the backward pass out of nine use both the tensor and vector pipelines. And these kernels account for 89.03\% of time in the forward pass and 91.87\% of the time in the backward pass. The SSM block contains operations (e.g., dot products) that use the tensor pipeline and element-wise multiplication, floating multiply and add and exponentials that use the vector pipeline and special function unit pipelines on the other hand leading to this behavior. Figure~\ref{fig:tensor2vector} shows the ratio between the tensor and vector FLOPS for such kernels. Modern day GPUs, such  as the H100 used in this study, have a higher number of tensor cores as opposed to vector units, especially when using lower precision tensor operations such as FP16 and FP8. For example on the H100, the ratio of the FP16 tensor cores to vector units is 16. This means anytime the tensor to vector FLOPS ratio of the workload is less than 16, the workload is bound by vector compute. If we pair that with what we observe in Figure~\ref{fig:tensor2vector}, the mixed tensor and vector compute kernels are almost entirely dominated by vector compute. This suggests that once issues such as the memory bandwidth problem are solved, we will be limited by vector compute. Thus, scaling the provisioned vector compute on the hardware is going to be critical. Further, software optimizations to effectively execute tensor and vector operations efficiently, such as proposed by Shah et al.~\cite{shah2024flashattention}, would be beneficial for SSM workloads to maximize utilization of both the tensor and vector computing pipelines.
\section{Conclusion}
In this work we analyzed the key properties of Mamba-based neural network models and profile the performance characteristics of these new models.  SSMs provide an alternative path to scaling performance compared to attention-based mechanisms. While matrix multiplication still consumes a significant runtime portion, SSM  block kernels also consume a significant amount of time.  We need new mechanisms to speed them using optimizations of both hardware and software, which will be key if we want to continue scaling the performance of Mamba based models.

\newcommand*\circled[1]{\protect\tikz[baseline=(char.base)]{
    \protect\node[shape=circle,draw,inner sep=1pt] (char) {#1};}}
\newcommand{\figref}[1]{Figure~\ref{#1}}
\newcommand{\tabref}[1]{Table~\ref{#1}}
\newcommand{\secref}[1]{Section~\ref{#1}}

\bibliographystyle{IEEEtran}
\bibliography{IEEEabrv,ref}

\end{document}